\documentclass[sigconf]{acmart}

\AtBeginDocument{%
  \providecommand\BibTeX{{%
    \normalfont B\kern-0.5em{\scshape i\kern-0.25em b}\kern-0.8em\TeX}}}

\acmConference[KDD '22]{KDD '22: ACM SIGKDD Conference on Knowledge Discovery and Data Mining}{August 14--18, 2021}{Washington, DC, USA}
  
\usepackage{subcaption}
\usepackage{graphicx}
\usepackage{tabularx}
\usepackage[export]{adjustbox}
\usepackage{diagbox}
\usepackage{rotating}
\usepackage{multirow}
\usepackage{slashbox}

\begin{document}

\title[Catalog Phrase Grounding (CPG): Grounding of Product Textual Attributes in Product Images]{Catalog Phrase Grounding (CPG):  Grounding of Product Textual Attributes in Product Images for e-commerce Vision-Language Applications}

\author{Wenyi Wu}
\affiliation{%
  \institution{Amazon}
  \streetaddress{550 Terry Ave N}
  \city{Seattle}
  \country{USA}}
\email{wenyiwu@amazon.com}

\author{Karim Bouyarmane}
\affiliation{%
  \institution{Amazon}
  \streetaddress{550 Terry Ave N}
  \city{Seattle}
  \country{USA}}
\email{bouykari@amazon.com}

\author{Ismail Tutar}
\affiliation{%
  \institution{Amazon}
  \streetaddress{550 Terry Ave N}
  \city{Seattle}
  \country{USA}}
\email{ismailt@amazon.com}

\renewcommand{\shortauthors}{Wu and Bouyarmane, et al.}

\begin{abstract}
We present Catalog Phrase Grounding (CPG), a model that can associate product textual data (title, brands) into corresponding re- gions of product images (isolated product region, brand logo region) for e-commerce vision-language applications. We use a state-of- the-art modulated multimodal transformer encoder-decoder archi- tecture unifying object detection and phrase-grounding. We train the model in self-supervised fashion with 2.3 million image-text pairs synthesized from an e-commerce site. The self-supervision data is annotated with high-confidence pseudo-labels generated with a combination of teacher models: a pre-trained general do- main phrase grounding model (e.g. MDETR) and a specialized logo detection model. This allows CPG, as a student model, to bene- fit from transfer knowledge from these base models combining general-domain knowledge and specialized knowledge. Beyond immediate catalog phrase grounding tasks, we can benefit from CPG representations by incorporating them as ML features into downstream catalog applications that require deep semantic under- standing of products. Our experiments on product-brand matching, a challenging e-commerce application, show that incorporating CPG representations into the existing production ensemble system leads to on average $5\%$ recall improvement across all countries globally (with the largest lift of $11\%$ in a single country) at fixed $95\%$ precision, outperforming other alternatives including a logo detection teacher model and ResNet50.
\end{abstract}

%

%
\keywords{Phrase Grounding, Object Detection, Transformers, MultiModal Model, Natural Language Understanding}

\maketitle

\section{Introduction}
Incorporating multimodal understanding of image and textual data of products is essential for many e-commerce applications. A typical product page on an e-commerce website consists of a product title in the form of short textual description, and an image representation of the product. The image can display an isolated product image or the image of the product in the context of use in an environment. Additional fields can be found in the product page, such as brand, dimensions, etc. An e-commerce catalog constitutes, therefore, a very rich corpus that can be used to design self-supervised vision-language tasks for the pre-training of deep learning product-vision-language understanding models. 

One such vision-language self-supervision task that can be crafted from the e-commerce catalog and that takes particular advantage of its specificities is phrase grounding\citep{chen2017query}. Phrase grounding consists in \emph{associating} (or \emph{grounding}) a textual phrase or part of it to a specific region of an image. The nature of e-commerce product data allows for expressive and domain-specific multi-task phrase grounding pre-training: grounding product title noun to isolated product region in the image, grounding product brand field to brand logo region in the product image, etc. We call the multi-task phrase grounding of e-commerce specific entities such as product-brand-to-logo and product-noun-to-object \emph{Catalog Phrase Grounding} (CPG). CPG outputs semantic rich representations that are particularly suited for e-commerce domain-specific downstream tasks.

\begin{figure}[b]
  \centering
  \includegraphics[width=1\linewidth,valign=t]{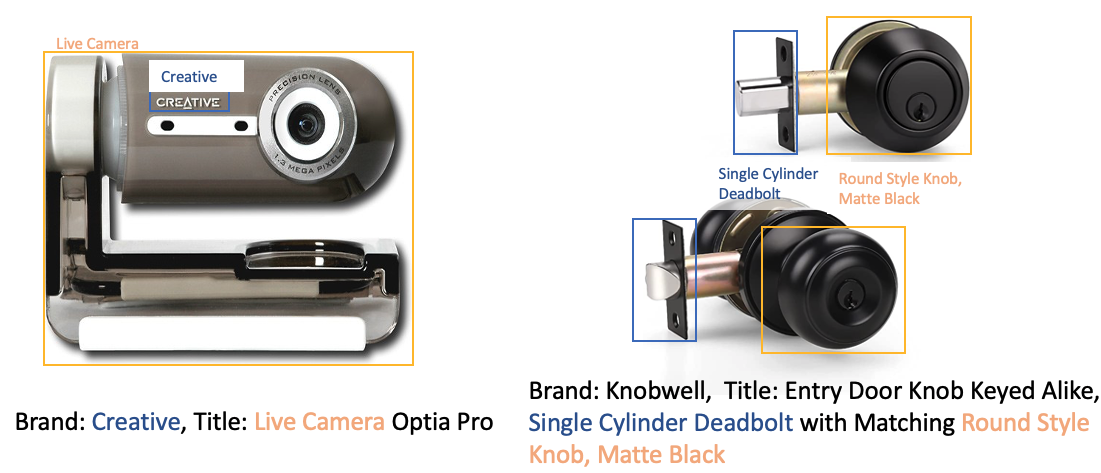}
  \caption{Left: logo localized by the logo detection teacher model and product localized by the phrase-grounding teacher models; Right: only product regions localized by the phrase-grounding teacher model. Teacher models can locate rare entities expressed in product titles.}
  \label{fig:annotation_exp}
\end{figure}

Pre-training of the CPG model is done in a self-supervised way with two teacher models: a pre-trained general-domain phrase grounding model, and a specialized logo-detection model. The student model learns to combine the knowledge distilled from both teacher models in multi-teacher multi-task learning setting. The tasks and phrases are crafted and self-generated from the catalog corpus, allowing to benefit from very large amount of unlabeled data. We illustrate two products with self-generated annotations in Figure \ref{fig:annotation_exp}.

The learned CPG visual representations are powerful, semantic-rich, fine-grained representations of the e-commerce product images. We demonstrate it with a challenging task: product-brand matching. A brand is a complex e-commerce entity that is represented by fields such as the brand name, brand logo (rarely), as well as a set of representative sample products. Brand name alone is not sufficient to characterize the brand entity, due to the sheer number of brands and homonyms. Matching a product to a brand entity thus consists in using the information from the brand and the representative sample products and inferring whether the query product \emph{belongs to} that brand. We show that GPG representations significantly improve the previous SOTA brand-matching system.

\section{Related Work and Contributions}

A standard joint visual and textual understanding model \citep{chen2020uniter, li2020unicoder} is typically trained with a fixed set of visual concepts (classes) on vision-language tasks, such as visual question answering \citep{antol2015vqa}, object detection \citep{zhao2019object} and phrase grounding. MDETR\citep{kamath2021mdetr} extended transformer based object detection model, i.e. DETR \citep{carion2020end}, to a modulated multimodal model trained with two tasks: object detection and phrase grounding. Therefore, MDETR could be pre-trained with $1.3$M text-images pairs having explicit alignment between phrases in text and objects in the image from combined pre-existing datasets, e.g. MS COCO\cite{lin2014microsoft}, Flickr30k \cite{plummer2015flickr30k} and etc. 

In order to further expand the visual concepts of image regions beyond vocabularies of pre-existing datasets, a recent line of work \cite{radford2021learning, li2021grounded, jia2021scaling} considers using web-scale raw image-text pairs.  CLIP \cite{radford2021learning} demonstrates that an image-level representations can be learned effectively through alignment between raw image-text pairs collected from the internet. Following the same idea, GLIP \cite{li2021grounded} pre-trains a phrase grounding model with $24$ million web-crawled image-text pairs. The regions of interest in images were detected by a pre-trained teacher model. It has been shown that the pre-trained GLIP learned effectively from the broader set of raw text to generate semantic enriched region-level visual representations. However, these models are not trained for e-commerce specific vocabulary and entities in synergy with downstream e-commerce applications.

Our proposal, CPG, is a transformer encoder-decoder architecture based model learning semantic rich visual representations for e-commerce specific entities through multiple domain-specific tasks. Following the trend of using free-form text, we train the CPG model with $2.3$M product entities synthesized from an e-commerce site in a self-supervised fashion. The bounding boxes for product-brand-to-logo grounding task are generated from a YOLO-based logo detection model \cite{fehervari2019scalable} while the product brands exist in product page already. The bounding boxes for product-noun-to-object task are generated by a pre-trained general domain modulated detection model\citep{kamath2021mdetr} conditional on noun phrases parsed from free-form product title using a general NLP parser\citep{bird2009natural}. 

\begin{figure}[b]
  \centering
  \includegraphics[width=1\linewidth,valign=t]{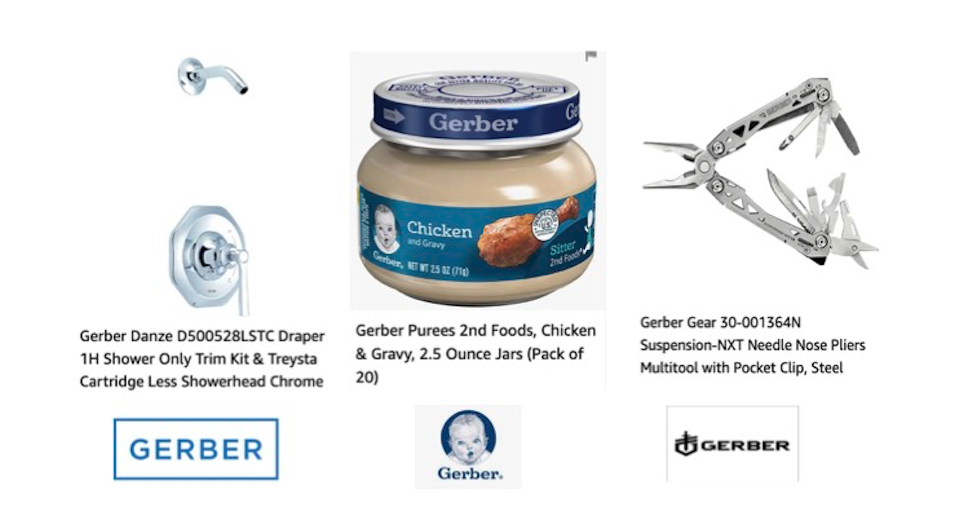}
  \caption{Homonym brands are independent brands associated with the same name (e.g. Gerber) selling different products. The \textit{Gerber} baby food has the logos in image while \textit{Gerber} tools and \textit{Gerber} plumbing fixtures don't.}
  \label{fig:homonyms}
\end{figure}

We present usage of the CPG representations as ML features to address one of major challenges of downstream product-brand matching application, which is differentiating homonym brands, especially when logos are absent, as shown in Figure \ref{fig:homonyms}. The GPG representations contain comprehensive visual-language understanding of logos, brand strings, product details for the query product entity and for all brand representative product entities. Therefore, the similarity between them shed light on identifying the correct brand of an input product from homonym ones, either through straightforward logo comparison or through product region comparison in the absence of logos.

We summarize the contributions of our work as follows:

\begin{itemize}
\item We propose an efficient and scalable method to learn semantic rich visual representations for e-commerce products in a self-supervised fashion. We leveraged massive raw product text data and images and applied teacher models to obtain region-phrase alignment annotations. In this way,  we extended the limited general vocabulary to substantial visual concepts expressed in the e-commerce catalog. 
\item  We transfer knowledge from 2 teacher models: a logo detection model and a general domain phrase grounding model by leveraging high-confidence predictions as pseudo labels for catalog specific tasks so that CPG benefits from both general-domain knowledge and specialized catalog knowledge.
\item We leverage the learned CPG representations to address challenging product-brand matching task and show improved performance. 
\end{itemize}

\section{Modal Architecture and Training}

\subsection{CPG Model Architecture}
\label{sec:model_arch}
\begin{figure*}
  \centering
  \includegraphics[width=1\linewidth,valign=t]{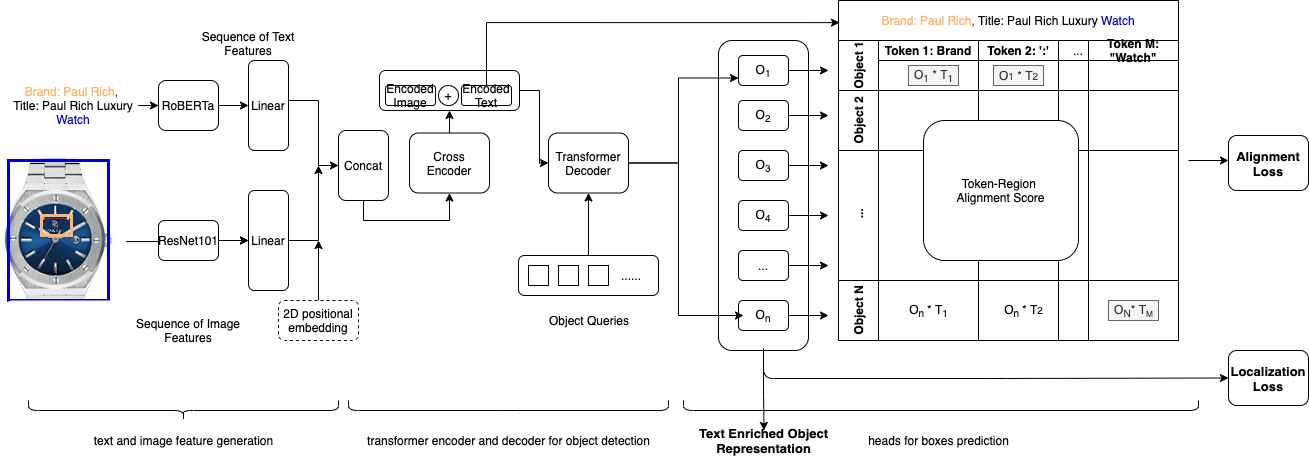}
  \caption{CPG has a unified framework for logo detection and product phrase grounding. CPG jointly train a text encoder, a image encoder, and a transformer-based region detection model (DETR) to predict the correct pairings of the mixture of (logo, brand) and (product, phrases) training examples and region locations simultaneously.}
  \label{fig:architecture}
\end{figure*}

The CPG model in this paper grounds logos and isolated product details to the brand and noun phrases in free-form title simultaneously. We manually craft the caption of image by concatenating the brand string and the product title together and inserting comma in between. To facilitate understanding of the logo concept and distinguishing noun words in brands from general vocabulary, e.g. brand \textit{Apple} versus fruit \textit{apple}, we prefix the brand string with "Brand :" and prefix the free-form title with "Title: ". Hence, given the product with brand "Paul Rich" and title "Paul Rich Luxury Watch", the crafted caption is "Brand: Paul Rich, Title: Paul Rich Luxury Watch". Given the product image and augmented caption, the CPG model is trained to ground logo regions to "Brand: Paul Rich" and to ground the watch region to "watch". We illustrate CPG model architecture in Figure \ref{fig:architecture}. 

The proposed CPG model architecture consists of 3 parts. First, the manually crafted image caption is tokenized and encoded using a pre-trained text encoder: RoBERTa\cite{liu2019roberta}. The text encoder has 12 transformer encoder layers, each with hidden dimension of 768 and 12 heads in the multihead attention. The product image is encoded using a pre-trained image encoder: ResNet101\cite{he2016deep}. Additionally, we concatenated encoded image vector with 2-dimensional positional embedding to conserve the spatial information. The encoded text and image are projected into a shared latent space by two independent linear projection functions. 

Second, the image and textual features are concatenated as a multimodal vector and fed to a joint transformer encoder with cross attention between image and textual features. Following DETR, we feed the model with a fixed small set of random initialized object queries (100), each of which is used to represent an object, i.e. a logo or a product region, in the image. CPG applies a transformer decoder to object queries while cross attending to the final hidden state of the cross encoder. Same as DETR, we use 6 encoder layers, 6 decoder layers and 8 attention heads in the attention layers. The learned object queries are textual-attribute-semantics enriched visual representations for logos and products, which are leveraged for downstream applications. 

Finally, CPG reasons the region locations and relations between phrases and regions simultaneously without differentiating whether the region is associated with a logo or a product detail. Each object representation is fed into two prediction heads: an token-region alignment head and a bounding box regression head, which are trained with contrastive alignment loss $\mathcal{L}_{align}$ and localization loss $\mathcal{L}_{loc}$, respectively. The CPG model aims to optimize the combined loss function as below.
\begin{equation}
\mathcal{L} = \mathcal{L}_{align} + \mathcal{L}_{loc}
\end{equation}

For contrastive alignment loss between noun phrases spanned multiple tokens and object queries, we use the same fine-grained contrastive loss proposed in MDETR\citep{kamath2021mdetr}. We consider the maximum number of tokens to be $M$, the number of the fixed set of object queries to be $N$, the set of tokens representing object $o_i$ to be $T_i^+$ and the set of objects associating with token $t_i$ to be $O_i^+$. The contrastive loss for objects is normalized by the number of positive tokens:
\begin{equation}
l_o = \sum_{i=0}^{N-1}\frac{1}{|T_i^+|}\sum_{j\in T_i^+}-log(\frac{exp(o_i^Tt_j/\tau)}{\sum_{k=0}^{M-1}\exp(o_i^Tt_k/\tau)})
\end{equation}
By symmetry, the contrastive loss for all tokens is given by:
 \begin{equation}
l_t = \sum_{i=0}^{M-1}\frac{1}{|O_i^+|}\sum_{j\in O_i^+}-log(\frac{exp(t_i^To_j/\tau)}{\sum_{k=0}^{N-1}\exp(t_i^To_k/\tau)})
\end{equation}
where $\tau$ is a temperature parameter set to 0.07. The average of $l_t$ and $l_o$ is used as contrastive alignment loss $\mathcal{L}_{align}$ to be minimized during model training.

\subsection{Scalable Training with Pseudo Labels}
In order to extend general vocabulary to noun phrases in product titles in e-commerce platform, the traditional way is to manually label phrases in corpus. Furthermore, to train a unified model for logo detection and product phrase grounding, it requires manual labels of bounding boxes which is expensive. Therefore, to scale the CPG model training with abundant textual concepts in e-commerce, we query $2.3$M raw product images and textual attributes from English countries of an e-commerce site and use them in a self-supervised framework. In order to generate product-noun-to-object grounding pseudo labels, we start with noun phrases extraction. Given augmented product image caption combining product brand and title attributes, as described in section \ref{sec:model_arch}, we apply general Natural Language Processing parser \citep{bird2009natural} to identify all noun words in title and include all leading adjective words prior to noun words to obtain noun phrases. We then apply a unified object detection and phrase grounding teacher model, MDETR\cite{kamath2021mdetr} to detects product regions conditioned on the extracted noun phrases. We obtained $3.2$M unique noun phrases from titles with $6.1$M associated bounding-boxes. To collect product-brand-to-logo grounding pseudo labels, we simply apply a logo detection model \cite{fehervari2019scalable} that is trained with e-commerce product images to localize logo regions in the image if exist. Then we associate logo regions to the brand section in augmented image captions. We obtained $92$k logo region grounding labels in total. We illustrate two sample annotations in Figure\ref{fig:annotation_exp}, the MDETR can localize rare language concepts expressed in the free-form product title, like \textit{camera}, \textit{single cylinder deadbolt}, and \textit{round style knob}. Finally, we train the CPG model on these two tasks simultaneously with two types of grounding labels together. 

Our self-supervised data augmentation is inspired by GLIP\cite{li2021grounded} which applied the pre-trained GLIP model to obtain pseudo-labels of raw web-crawled image-text pairs to finetune the same GLIP model. We first collect samples from an e-commerce site to scale grounding data and especially to enrich e-commerce related semantic concepts which is different from general domain. We step further to obtain pseudo-labels from two pre-trained models so that as a student model, CPG model benefits from both general knowledge transferred from MDETR and specialized brand knowledge transferred from the logo detection model. Because of cross attention between logos and product details, CPG model enables product understanding in both the general context and the brand specific context. Furthermore, CPG model enables comprehensive brand understanding by unifying logo understanding and representative sample products understanding. Therefore, CPG model outperforms the logo detection teacher model when applying to the product-brand matching task, which will be described later in section \ref{sec:product_brand}.

\section{Product-Brand Matching Application}
\label{sec:product_brand}
We show the effectiveness of CPG representations pre-trained with pseudo labels by applying them to a downstream e-commerce application: product-brand matching. Brand is a key attribute impacting customers' shopping decisions, which, however, is not trivial to infer because of homonyms. Differentiating homonym brands is challenging and can only be done if different logos and/or different products sold by brands are provided. For example, we should map a\textit{Gerber} knife to the right \textit{Gerber} in Figure \ref{fig:homonyms} instead of the other two, because we learn from the representative product that right \textit{Gerber} sells gears while the middle one sells baby food and the left one sells plumbing fixtures. For the middle \textit{Gerber}, we can easily map products correctly by identifying the logo. Therefore, in order to map products to brands correctly, we have to comprehensively understand the brand entity consisting of multiple products and logos (may not exist). The number of brand representative products varies per brand because of remarkable brand scope difference. Some brands, like \textit{3M},  sell products from dozens of different categories while others focus on one specific category. We formulate the product to brand matching problem as an asymmetric entity matching problem, which makes binary predictions regarding whether a product belongs to a brand. The existing state-of-the-art product-brand matching system ensembles a wide variety of features based on product text data (details in \ref{sec:base_model}). It fails to capture image features and hence is incapable of distinguishing homonym brands by logos or other fine-grained regions in product images. 

We use the semantics rich CPG representations extracted from the model with high confidence ($>0.5$) as additional ML features to the existing system. We can not simply concatenate CPG representations and feed to downstream task because the number of representative products per brand has a long tailed distribution. The number of object representations with high confidence also varies from one image to another. As a result, concatenating them and padding to the longest makes the feature vector extraordinary long, which increases the matching system complexity and therefore breaks the latency requirement for real-time use cases. Another way is to truncate CPG representations to a fix number. However, it will prevent model from making informative predictions. For example, if \textit{Apple} has only one representative product, e.g. a computer, model can not tell if \textit{Apple} phone case should be mapped to it. Therefore, in order to utilize all information but still have a light enough system to satisfy latency requirement, we first compute distance-based similarity scores, i.e. Euclidean distance and Cosine distance, between CPG representations of the input product and CPG representations of each representative product. Then the summary statistics, i.e. minimum, maximum, medium and variance, of similarity scores are fed to ensemble system. In addition, we feed two boolean values to indicate whether the number of learned CPG representations is $0$ for the product and for the brand, respectively. We refer these summary statistics derived from CPG representations as CPG features. 

\section{Experimentation and Results}
We first train the CPG model using augmented product image text pairs with pseudo labels. Then we evaluate the textual-attribute-semantic rich object representations learned by CPG by supplementing them to existing product-brand matching model. We report the relative gain of recall at $95\%$ precision in 9 countries as we need to meet a high precision threshold for deployment to ensure customer shopping experience.  We further compare the visual representations learned by CPG with representations learned from two types of vision model: the logo detection model and image-level understanding model, i.e. ResNet50. 

\subsection{Product-Brand Matching Dataset}
All samples for product-brand matching task are in (product, brand) pair format. Input product is a structured entities with a fixed set of attributes, e.g. title, brand and image. The textual attributes could be inaccurate or even missing. Input brand is also a structured entities with varying number of representative products that share the same data structure as the input product. 
We collected $50, 000$ (product, brand) training pairs per country from 6 countries (we will refer them as A-F). The textual attributes in these countries are in English and multiple Romance languages, e.g. French, Spanish and etc. The training set are randomly split into $80\%$ training set and $20\%$ validation set. For test, we collected $20,000$ (product, brand) pairs per country from 9 countries globally consisting of A-F and 3 new English countries (we will refer to them as G, H and I). We use different sampling strategies to collect training and test samples. For training, in order to utilize auditing resources efficiently, we exclude trivial negative pairs, the products in which are obvious generic products irrelevant to any known brand entity. To collect test pairs, we simply random sample from the catalog of an e-commerce site. The catalogs in different countries show different patterns. In country I, there are more products with homonym brands than the rest and therefore, it is expected to show performance lift once we include image signals. We denote a pair as positive if the product is sold by the brand and as negative if not. We collect ground truth of product and brand relationships through manual labeling.

\subsection{Baseline Models}
\subsubsection{Existing ensemble model with Text Features}
\label{sec:base_model}
The existing product-brand matching model ensembles in a total of 139 features containing manually crafted syntactic similarity features and ML features learned from base models, e.g. brand extraction model, textual attributes understanding model and etc. The brand extraction model tackles the challenge of missing brand attribute  when brand strings are mentioned in other textual attributes. The textual attributes understanding model captures semantic similarity between the product and the brand despite the brand string variation, e.g. similarity between \textit{James Bond 007 Fragrances} and \textit{007 Frangrances}. This existing ensemble model doesn't contain any signal from product image, which serves as the first baseline to the ensemble model with features derived from object representations learned from CPG. The existing model is a country-aware model trained with all training samples collected from 6 countries. The performance of the ensemble model is evaluated in each of 9 countries. 

\subsubsection{Ensemble model with logo features}
\label{sec:logo_model}
One natural way to distinguish homonym brands is to compare logos. We apply the YOLO-based logo detection teacher model to both input product image and brand representative product images and use detected logos as features. Because the number of representative products per brand varies, directly using detected logos leads to various input length. Furthermore, the detected logo region similarity could be impacted by the original product image quality, size, angle and etc. To address these challenges, we first leverage encoded vector from the last hidden layer of the logo detection model to replace raw detected logo regions as input features. Then, to address various input length challenge and have fair comparison with CPG representations, we supplement the existing ensemble model with the same set of summary statistics based on similarity between detected logos, as described in \ref{sec:product_brand}.  We refer this set of features as logo features. 

\subsubsection{Ensemble model with ResNet50 features}
\label{sec:resnet_model}
Another way to leverage image information is through image-level understanding. We fine-tuned a ResNet50 model using product images synthesized from English e-commerce catalogs. The data collection process is independent of CPG self-supervised training data collection process. The fine-tune task is to detect whether images are from duplicate products. The training samples contains similar yet different products whose images differ in details as well as duplicate products whose images may be taken from different angles, lights and etc. Therefore, the fine-tuned ResNet50 model learns e-commerce specialized image patterns and learns to pay attention to both local regions as well as the whole image. We use vectors from last hidden layer as an image-level representations for product image. The same set of summary statistics of Euclidean and Cosine distances between image representations of the input product and of the brand representative products are supplemented to existing ensemble model for re-training and evaluation. We refer this set of features as ResNet50 features

\subsection{Results}
We note that all results reported in this paper are in absolute terms. We evaluate model performance using recall at high precision because we need to prevent incorrect brand mappings from impairing customers' shopping experience. Therefore, we report relative gains in terms of recall at precision $90\%$ and recall at precision $95\%$ and denote lift as  $\Delta R@P90$ and $\Delta R@P95$ in tables.


\subsubsection{Comparison with Existing Ensemble model with text features}
Table \ref{tab:base_model} shows the performance lift of the ensemble model with CPG features over the existing ensemble model with text features only as described in \ref{sec:base_model} across 9 countries. From the table, we see that leveraging similarity between object representations learned from CPG model leads to significant performance jump on top of existing ensemble model in all countries globally except one country C. In country C, the model with CPG features performs comparably with the baseline model because textual attributes quality is high in country C where we observe less inaccurate or missing attributes. Therefore, the model with text features already contain enough product information to distinguish homonym brands by understanding the textual attributes. As expected, we observe the largest performance lift in country I with largest homonym proportion. This confirmed the conjecture that semantic rich object representations provide signals to differentiate homonym brands.

\begin{table*}
\centering
\begin{tabular}{c|ccccccccc}
\toprule
 & A & B & C & D & E & F & G & H & I \\
\midrule
$\Delta$ R@P90 & 1.8\%   & 0.1\%  & 0.0\%  & 0.0\%  & 1.5\%   & 0.0\%  & 2.4\%  & 8.2\%  & 9.1\%  \\
$\Delta$ R@P95 & 3.9\%  & 5.0\%  & 0.0\%  & 4.0\%  & 4.5\%  & 5.6\%  & 7.6\%  & 7.1\%  & \textbf{11.4\%}\\
\bottomrule 
\end{tabular}
\caption{ Performance gains of supplementing CPG features to the existing ensemble model with features derived from textual attributes only in 9 countries}
\label{tab:base_model}
\end{table*}

\begin{table*}
\centering
\begin{tabular}{c|ccccccccc}
\toprule
 & A & B & C & D & E & F & G & H & I \\
\midrule
$\Delta$ R@P90 & 1.7\%   & 0.1\%  & 0.0\%  & 0.0\%  & 2.9\%   & 1.4\%  & 0.9\%  & 9.0\%  & 8.0\%  \\
$\Delta$ R@P95 & 1.6\%  & 4.0\%  & 0.0\%  & 0.0\%  & 7.3\%  & 6.9\%  & 4.6\%  & \textbf{13.3\% }  & 3.8\%\\
\bottomrule
\end{tabular}
\caption{ Performance gains of leveraging CPG features over leveraging features derived from the logo detection teacher model in 9 countries}
\label{tab:logo_model}
\end{table*}

\begin{table*}
\centering
\begin{tabular}{c|ccccccccc}
\toprule
 & A & B & C & D & E & F & G & H & I \\
\midrule
$\Delta$ R@P90 & 0.6\%   & 0.0\%  & 0.0\%  & 0.0\%  & 2.9\%   & 0.0\%  & -0.4\%  & 5.7\%  & 8.7\%  \\
$\Delta$ R@P95 & 4.1\%  & 3.8\%  & 0.0\%  & 2.0\%  & 2.8\%  & 2.8\%  & 5.2\%  & 4.8\%  & \textbf{10.5\%}\\
\bottomrule 
\end{tabular}
\caption{ Performance gains of leveraging CPG features over leveraging features derived from the ResNet50 model fine-tuned on catalog from English countries for image matching in 9 countries.}
\label{tab:resnet_model}
\end{table*}

\subsubsection{Comparison with Ensemble model with logo features}
Table \ref{tab:logo_model} shows the performance lift of the ensemble model with CPG features over the ensemble model with logo features as described in \ref{sec:logo_model} across 9 countries. We can see that, leveraging object representations learned from CPG model performs better than simply using detected logos from the teacher model in all countries except C and D where these two models perform comparably. Despite the performance lift in country I in table \ref{tab:logo_model} is lower than in table \ref{tab:base_model}, we still observe positive performance lift. This indicates that leveraging logo features can only mitigate the homonym challenge partially. It doesn't solve all problems because not all product images contain the logo. In fact, only $40\%$ of product images have logo detected with high confidence (>$0.5$). For cases where the input product image doesn't contain logos, the object representations learned from CPG can provide fine-grained product image understanding to guide the correct brand mapping.

\subsubsection{Comparison with Ensemble model with ResNet50 features}
Table \ref{tab:resnet_model} shows the performance lift of the ensemble model with CPG features over the ensemble model with image-level understanding features learned from the ResNet50 model as described in \ref{sec:resnet_model} across 9 countries. We see that CPG features lead to higher performance gains in all countries except country C and G. In country G, both models perform comparably at precision $90\%$ and our model performs better at precision $95\%$.  All ensemble models perform similarly in country C. The lift of leveraging CPG features on top of leveraging ResNet50 features are mainly from two sources: more fine-grained image understanding and logo specialized knowledge. In country I, where logo features lead to performance lift, we see that CPG features outperform ResNet50 features significantly. This demonstrates that CPG model effectively transfers knowledge from the logo detection teacher model while ResNet50 features are lack of logo specialized knowledge.

\section{Conclusion}
In this paper, we presented a \emph{catalog phrase grounding} (CPG) model which learns fine-grained visual representations of product image conditional on structured product textual attributes. We investigated how to train the model with catalog-scale raw product attributes in a self-supervised fashion by transferring knowledge from both the general domain phrase grounding teacher model and the catalog specific logo detection teacher model. Therefore, the learned representations of logos and isolated product details from the same latent space provide an integrated understanding of brands and products. The effectiveness of learned semantic rich representations are demonstrated by performance gains on a crucial and challenging e-commerce application: product-brand matching. We further show that integrated understanding of products and brands makes CPG competitive with the logo detection teacher model. It's worth future study to apply CPG representations to other e-commerce applications.

\bibliographystyle{ACM-Reference-Format}
\bibliography{references}

\end{document}